\documentclass[10pt,twocolumn,letterpaper]{article}

\usepackage[pagenumbers]{cvpr} 

\usepackage{graphicx}
\usepackage{capt-of}
\usepackage{amsmath,amssymb,amsfonts,dsfont,bm,bbm,mathrsfs,pifont}
\usepackage{booktabs,multirow,adjustbox}
\usepackage[pagebackref,breaklinks,colorlinks]{hyperref}
\usepackage[capitalize]{cleveref}  
\crefname{section}{Sec.}{Secs.}
\Crefname{section}{Section}{Sections}
\crefname{table}{Tab.}{Tabs.}
\Crefname{table}{Table}{Tables}
\crefname{figure}{Fig.}{Figs.}
\Crefname{figure}{Figure}{Figures}
\crefname{equation}{Eq.}{Eqs.}
\Crefname{equation}{Equation}{Equations}
\usepackage{tabulary}
\newcolumntype{x}[1]{>{\centering\arraybackslash}p{#1pt}}
\newlength\savewidth

\makeatletter\renewcommand\paragraph{\@startsection{paragraph}{4}{\z@}
  {.5em \@plus1ex \@minus.2ex}{-.5em}{\normalfont\normalsize\bfseries}}\makeatother

\def\confName{CVPR}
\def\confYear{2022}

\begin{document}

\title{Cross-Model Pseudo-Labeling for Semi-Supervised Action Recognition}

\author{Yinghao Xu$^{1}$ \quad Fangyun Wei$^{3}$ \quad Xiao Sun$^{3}$ \quad Ceyuan Yang$^{1}$ \\ Yujun Shen$^1$ \quad Bo Dai$^2$ \quad Bolei Zhou$^1$ \quad Stephen Lin$^{3}$\vspace{3pt}\\
	\hspace{-35pt} $^1$The Chinese University of Hong Kong \quad
    $^2$S-Lab, Nanyang Technological University  \quad 
    $^3$Microsoft Research Asia \\
    \hspace{-30pt} {\tt\small \{xy119,yc019,sy116,bzhou\}@ie.cuhk.edu.hk \quad  bo.dai@ntu.edu.sg \quad } 
	{\tt\small \{fawe,xias,stevelin\}@microsoft.com} \\
}

\maketitle

\begin{abstract}
Semi-supervised action recognition is a challenging but important task due to the high cost of data annotation.
A common approach to this problem is to assign unlabeled data with pseudo-labels, which are then used as additional supervision in training.
Typically in recent work, the pseudo-labels are obtained by training a model on the labeled data, and then using confident predictions from the model to teach itself.
In this work, we propose a more effective pseudo-labeling scheme, called Cross-Model Pseudo-Labeling (CMPL).
Concretely, we introduce a lightweight auxiliary network in addition to the primary backbone, and ask them to predict pseudo-labels for each other.
We observe that, due to their different structural biases, these two models tend to learn complementary representations from the same video clips.
Each model can thus benefit from its counterpart by utilizing cross-model predictions as supervision.
Experiments on different data partition protocols demonstrate the significant improvement of our framework over existing alternatives.
For example, CMPL achieves $17.6\%$ and $25.1\%$ Top-1 accuracy on Kinetics-400 and UCF-101 using only the RGB modality and $1\%$ labeled data, outperforming our baseline model, FixMatch~\cite{sohn2020fixmatch}, by $9.0\%$ and $10.3\%$, respectively.
\footnote{Project page is at \url{https://justimyhxu.github.io/projects/cmpl/}.}
\end{abstract}

\section{Introduction}\label{sec:intro}

The rapid development of deep learning has led to great success in action recognition.
In the standard supervised learning protocol, a considerable number of annotated videos is needed but difficult to acquire in practice.
On the other hand, about 500 hours of video is uploaded to YouTube every minute worldwide, providing a tremendous amount of unlabeled data.
Leveraging such unlabeled videos for semi-supervised learning could thus be of great benefit for action recognition.

\begin{figure}[t]
    \centering
    \includegraphics[width=0.95\linewidth]{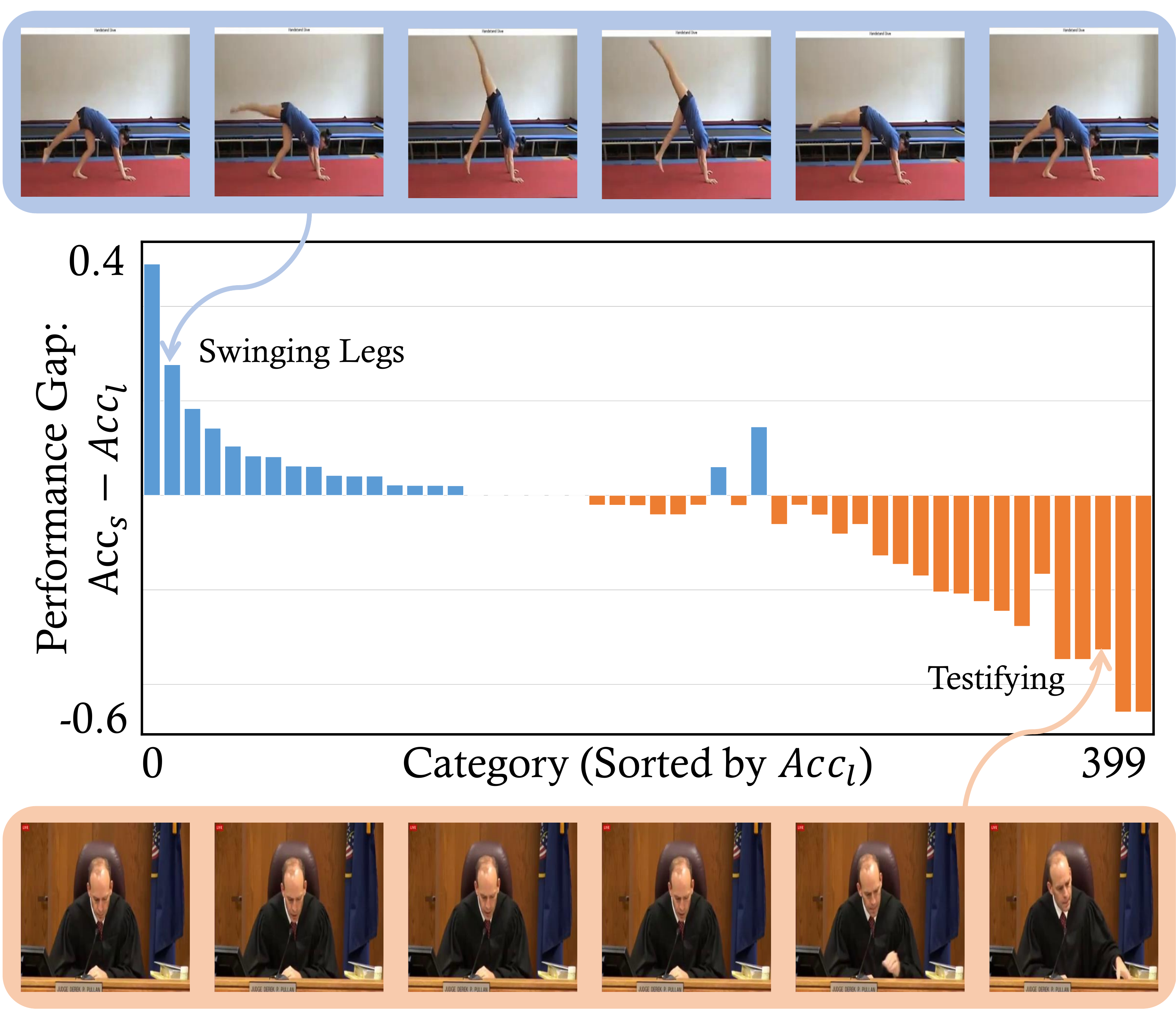}
    \vspace{-10pt}
    \caption{
        \textbf{Category-wise performance gap between small and large networks} under the \textit{supervised} training setting given $1\%$ labeled videos in Kinetics-400~\cite{kinetics}.
        $Acc_s$ and $Acc_l$ denote the accuracy of the small (3D-ResNet50$\times$1/4) and the large (3D-ResNet50) network, respectively.
        For categories on which the large network performs poorly (\textit{i.e.}, the left half of the figure), the small network behaves better even with much lower model capacity. 
        Concretely, the small network tends to perform well on classes with stronger temporal dynamics, \textit{i.e.}, ``Swinging Legs'', while the large network better recognizes actions mainly characterized by spatial information, \textit{i.e.}, ``Testifying''.
        See \cref{sec-experiment-empirical-learned} for further discussion.
    }
    \label{fig-observation}
    \vspace{-15pt}
\end{figure}

To gain supervision from unlabeled data, a common practice is to assign pseudo-labels to these data and treat them as ``ground-truth'' for training~\cite{lee2013pseudolabel, tarvainen2017meanteacher, xie2020selftraining, sohn2020fixmatch}.
Specifically, existing approaches train a model on labeled data and then use it to predict the unlabeled videos. If the confidence in a prediction is high enough, the prediction will be taken as the pseudo-label of the video, to be used in network training henceforth. The quantity and quality of pseudo-labels therefore has a significant impact in the current learning scheme. However, the limited discriminative power derived from a small amount of labeled data leads to inadequate pseudo-labels and limits the gain from unlabeled data.

To better capitalize on unlabeled videos, we present a pseudo-labeling approach based on complementary representations at the model level. 
We observe that models of different scales exhibit markedly different behaviors in regards to category-wise performance, due to their different structural biases.
As shown in \cref{fig-observation} and studied in \cref{sec-experiment-empirical}, a small model, despite its lower capacity, can achieve notable improvements over a large model on certain categories. In particular, it better captures temporal dynamics in recognizing actions, while a large model tends to better learn spatial semantics for distinguishing different action instances.
This indicates that the two models differ in what they learn and therefore can complement each other in pseudo-labeling, where one model can more successfully generate pseudo-labels for some categories while the second model can be more effective on others.

\begin{figure}[t]
    \centering
    \includegraphics[width=0.95\linewidth]{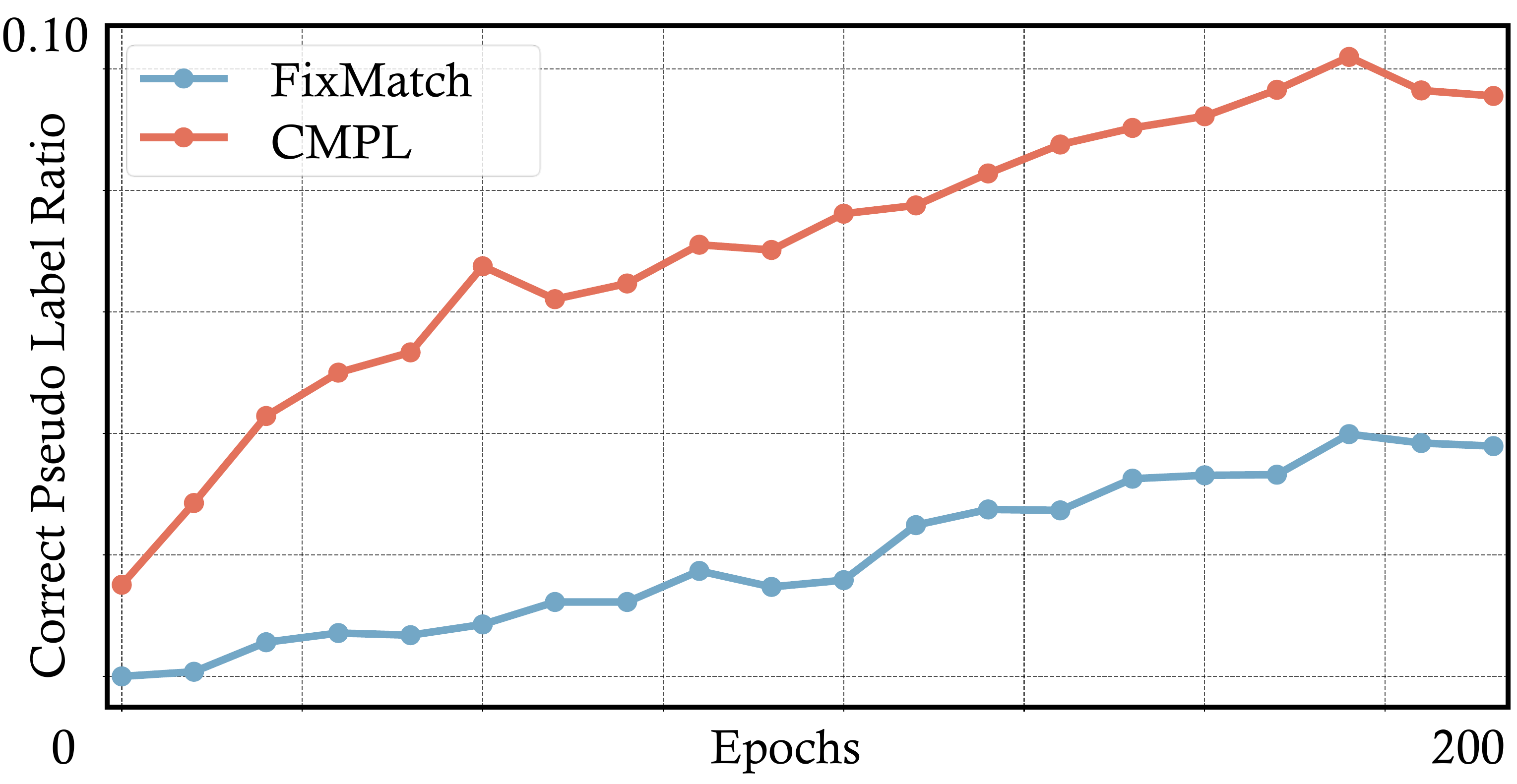}
    \vspace{-10pt}
    \caption{
        \textbf{Ratio of correct pseudo-labels}. To compare the quality and quantity of pseudo labels generated by CMPL and FixMatch, we plot the ratio of correct pseudo-labels relative to the total number of videos in the unlabeled dataset. Both models are trained with 1\% labeled Kinetics-400\cite{kinetics} (the others are unlabeled). Our CMPL is shown to more effectively provide high-quality pseudo-labels than FixMatch.
    }
    \label{fig-plratio}
    \vspace{-15pt}
\end{figure}

Based on this observation, we propose \textit{Cross-Model Pseudo-Labeling (CMPL)}, where the primary backbone is supplemented by a lightweight auxiliary network with a different structure and fewer channels than the backbone.
This difference in architecture leads to a different representation of the input data that complements that of the primary backbone.
Then, given an unlabeled video clip, we borrow a confident prediction from the auxiliary network as the pseudo-label for the primary backbone, and vice versa.
As these two models have their own strengths, a greater number of unlabeled videos can be engaged in pseudo-labeling, facilitating each other's training accordingly. \cref{fig-plratio} suggests that our CMPL obtains more high-quality pseudo-labels than the baseline. 
In addition, we also study the compatibility of this cross-model framework with conventional temporal data augmentations (\textit{i.e.}, adjusting the temporal location and the frame rate), which are widely used in recent literature~\cite{tcl, vthcl, feichtenhofer2021large}. 
Experiments on a range of standard benchmarks and training settings demonstrate the effectiveness of our CMPL.
In particular, when using only the RGB modality and $1\%$ labeled videos, CMPL achieves $17.6\%$ and $25.1\%$ Top-1 accuracy on Kinetics-400~\cite{kinetics} and UCF-101~\cite{soomro2012ucf101}, surpassing FixMatch~\cite{sohn2020fixmatch}, by $9.0\%$ and $10.3\%$, respectively.
We also conduct a comprehensive empirical analysis to study how the cross-model supervision helps improve performance. 
This analysis shows that the primary backbone has large improvement on classes for which the auxiliary network works very well, supporting our motivation that the auxiliary network can complement the backbone.

\section{Related Work}\label{sec-related}

\noindent{\textbf{Semi-Supervised Learning for Image Classification.}}
Semi-supervised learning has been well-explored in the area of image classification. Some prior works leverage consistency regularization which requires models to be robust to perturbations including data augmentations \cite{uda} and adversarial perturbations \cite{miyato2018vta}. Recent work focuses more attention on the pseudo-labeling framework, which assigns labels to unlabeled images according to model predictions. In particular, \cite{tarvainen2017meanteacher} and \cite{lee2013pseudolabel} produce pseudo-labels by utilizing the exponential moving average of model parameters and historical predictions, respectively. Differently, FixMatch~\cite{sohn2020fixmatch} combines consistency regularization and pseudo-labeling by requiring the predictions from strongly-augmented data to mirror those from weakly-augmented data. 
These methods do not explicitly consider the temporal dynamics that characterize human actions, which are shown in our work to be more effectively represented with the help of a separate complementary network.

\noindent{\textbf{Semi-Supervised Learning for Action Recognition.}}
While action recognition~\cite{twostream, tsfusion, cts,tsn,trn,tsm, r21d_v2, c3d, kinetics, p3d, slowfast, tpn, trajconv, trajpool, improvedtraj, nonlocal} has progressed significantly in recent years,
semi-supervised learning for action recognition has been less studied. \cite{iosifidis2014semi} represents actions by a conventional action bank and applies variants of extreme learning machines for final class prediction.
\cite{zeng2017semi} presents an encoder-decoder structure that is trained via an image reconstruction pretext task.
\cite{Jing_2021_WACV} introduces a new framework that leverages a 2D image classifier to assist action recognition.
\cite{tcl} proposes a temporal contrastive learning framework to model temporal aspects by comparing the same video at different speeds. 

A different approach is to learn from multiple views of the same source data. In the concurrent work of \cite{xiong2021multiview}, the different views correspond to RGB, optical flow, and temporal gradients. Prediction results are separately generated from each view using a single common model, and these results are ensembled to produce pseudo-labels for retraining the model.
This approach is similar to that of co-training~\cite{cotraining}, where two networks are asked to iteratively make predictions on the unlabeled data and the pseudo-labels are then merged to supervise the two models jointly.
Differently, our approach requires the two models with different architectures to provide pseudo-labels for each other in a cross-teaching manner.
In this way, they are expected to benefit from the complementary representation learned by their counterparts.

\section{Method}\label{sec:method}

\begin{figure*}[t]
    \centering
    \includegraphics[width=0.9\textwidth]{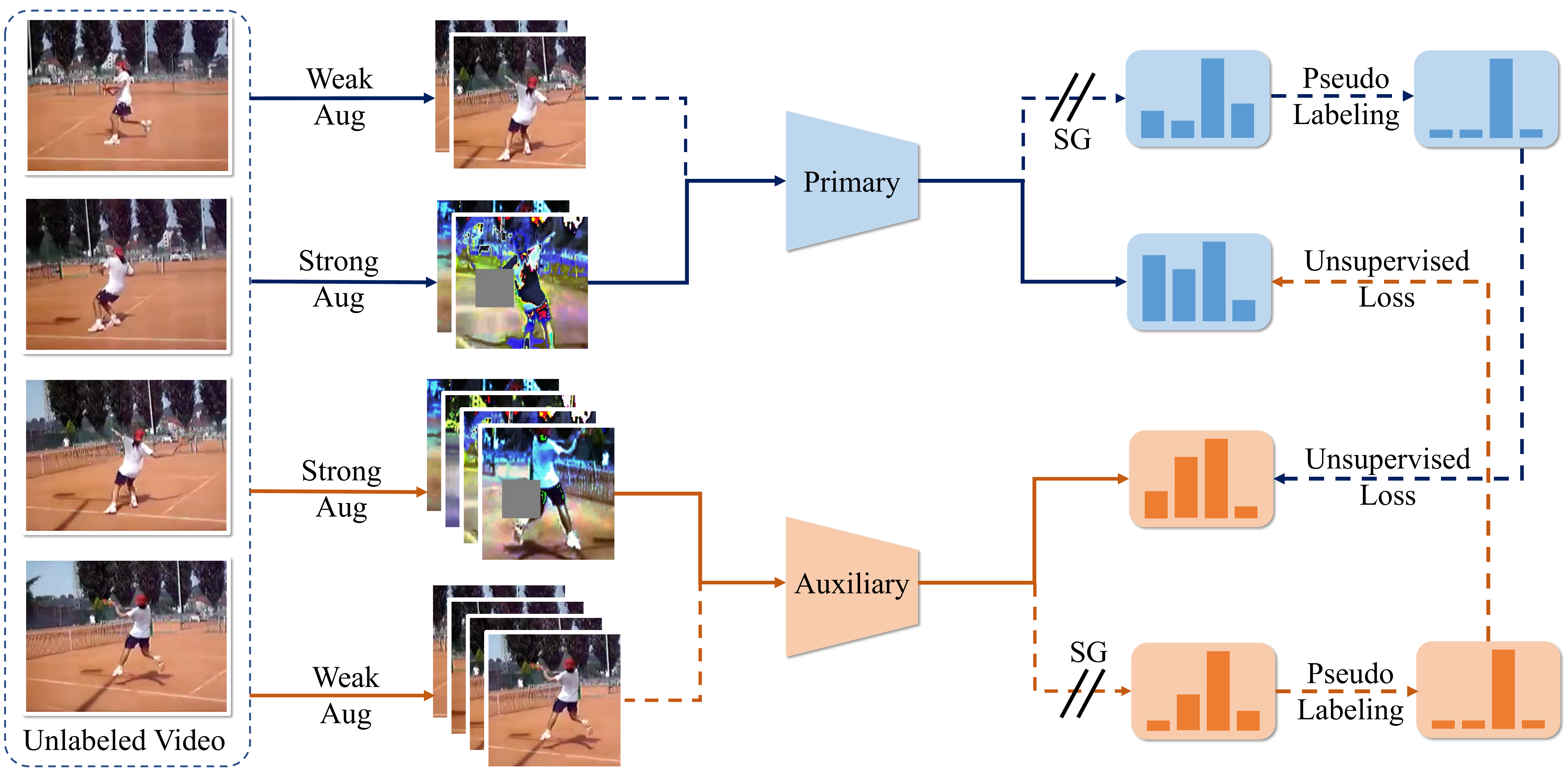}
    \vspace{-5pt}
    \caption{
        \textbf{Illustration of the proposed Cross-Model Pseudo-Labeling (CMPL) framework}, which consists of two models, \textit{i.e.}, the primary backbone $F(\cdot)$ and the auxiliary network $A(\cdot)$.
        They take video input at different frame rates.
        Unless specified, we adopt 3D-ResNet50 as the primary backbone and a lightweight 3D-ResNet50 $\times1/4$ as the auxiliary network.
        Given an unlabeled video, these two models make independent predictions on weakly-augmented data.
        The predicted results are used to produce a pseudo-label for their sibling where it is used as supervision for a strongly-augmented version.
        ``SG'' stands for the stop-gradient operation.
        The supervised losses from labeled data are omitted in this figure.
    }
    \label{fig-framework}
    \vspace{-5pt}
\end{figure*}

Given a labeled set
$\mathcal{V} = \{(v_1, y_1), \ldots, (v_{N_l}, y_{N_l})\}$ containing $N_l$ videos and an unlabeled set $\mathcal{U} = \{u_1, \ldots, u_{N_u}\}$ comprising $N_u$ videos, CMPL learns an action recognition model by efficiently leveraging both labeled and unlabeled data. In general, $N_u$ is much larger than $N_l$.
In this section, we first present preliminary background on the pseudo-labeling framework in \cref{sec-method-preliminary}, then we introduce the proposed CMPL framework in \cref{sec-method-cmpl}, followed by a description of the implementation for CMPL in \cref{sec-method-implementation}.

\subsection{Preliminaries on Pseudo-Labeling} \label{sec-method-preliminary}
Pseudo-labeling is a commonly used technique for semi-supervised image recognition.
It promotes the idea of utilizing the model itself to produce artificial labels for unlabeled data.
The artificial labels obtained with confidence above a pre-defined threshold are retained, and the corresponding unlabeled data can then serve as additional training samples.
A recent state-of-the-art method that adopts the pseudo-labeling scheme is FixMatch \cite{sohn2020fixmatch}, which uses a weakly-augmented image as input to acquire the image's pseudo-label, that is subsequently paired with a strongly-augmented version to form a labeled sample. 
FixMatch can be directly extended to semi-supervised action recognition as:
\begin{align}
   \mathcal{L}_{u} = \frac{1}{B_u} \sum_{i=1}^{B_u} \mathbbm{1} (\max (p_i) \geq \tau) \mathcal{H}(\hat{y}_i, F(\mathcal{T}_{strong}(u_i))), \label{eqa-fixmath}
\end{align}
where $B_u$ represents the batch size,
$\tau$ is a threshold to determine whether a prediction is confident or not, $\mathbbm{1}(\cdot)$ denotes the indicator function, 
$p_i = F(\mathcal{T}_{weak}(u_i))$ denotes class distribution, and $\hat{y}_i=\arg\max(p_i)$ represents the pseudo-label. 
$\mathcal{T}_{strong}(\cdot)$ and $\mathcal{T}_{weak}(\cdot)$ respectively denote the weak and the strong augmentation processes.
$\mathcal{H(\cdot, \cdot)}$ denotes the standard cross-entropy loss. 
$\mathcal{L}_u$ is the loss on the unlabeled data, and the loss on the labeled data is the cross-entropy loss commonly used in action recognition.

\subsection{Cross-Model Pseudo-Labeling}\label{sec-method-cmpl}

As described in \cref{sec-method-preliminary}, the core idea of recent semi-supervised learning approaches is to construct as many high-quality pseudo-labels for unlabeled data as possible. However, when the number of labeled data is limited, the discriminative power of a single model is too weak to successfully assign a large number of pseudo labels to the unlabeled data. 
Therefore, our approach is to learn two models with different architectures and ask them to provide pseudo-labels for each other,
which is inspired by the observation that different models have different structural biases that lead to complementary semantic representations.
As shown in \cref{fig-framework}, our CMPL framework employs two models (\textit{i.e.}, the primary backbone $F(\cdot)$ and the auxiliary network $A(\cdot)$) with different capacities. 
They both learn from labeled data through supervised training and simultaneously provide their companion with pseudo-labels for unlabeled data.
This symmetric design incentivizes the two models to learn complementary representations, which in turn boosts performance.

\noindent\textbf{Learning on Labeled Data.}
It is straightforward to train a model on labeled data. Given a set of labeled videos $\{(v_i, y_i)\}_{i=1}^{B_l}$, we formulate the supervised loss of the two networks as:
\begin{align} 
    \mathcal{L}_{s}^F = \frac{1}{B_l} \sum_{i=1}^{B_l} \mathcal{H}(y_i, F(\mathcal{T}_{standard}^F(v_i))), \label{eqa-supf} \\
    \mathcal{L}_{s}^A = \frac{1}{B_l} \sum_{i=1}^{B_l} \mathcal{H}(y_i, A(\mathcal{T}_{standard}^A(v_i))), \label{eqa-supa}
\end{align}
where $\mathcal{T}_{standard}(\cdot)$ represents the standard data augmentation used in action recognition~\cite{nonlocal, slowfast}.

\noindent\textbf{Learning on Unlabeled Data.}
Given an unlabeled video $u_i$, the auxiliary network $A(\cdot)$ will make a prediction on the weakly-augmented $u_i$ and output the category-wise probabilities, $p_i^A = A(\mathcal{T}_{weak}(u_i))$.
If the maximum entry among these probabilities, $\max(p_i^A)$, surpasses a pre-defined threshold $\tau$, 
we then regard this to be a solid prediction and 
utilize $p_i^A$ to derive the pseudo ground-truth $\hat{y}_i^A=\arg\max(p_i^A)$ for the strongly-augmented $u_i$.
In this way, the backbone $F(\cdot)$ can be learned with
\begin{align}
    \mathcal{L}_{u}^F=\frac{1}{B_u} \sum_{i=1}^{B_u} \mathbbm{1} (\max (p_i^A) \geq \tau) \mathcal{H}(\hat{y}_i^A, F(\mathcal{T}_{strong}(u_i))), \label{eqa-unsupf}
\end{align}
where $B_u$ denotes batch size, and $\mathcal{H(\cdot, \cdot)}$ is cross-entropy loss.

Similarly, the primary backbone will also make a prediction $p_i^F = F(\mathcal{T}_{weak}(u_i))$ and use it to generate a labeled pair $(\hat{y}_i^F
, \mathcal{T}_{strong}(u_i))$ for the auxiliary network:
\begin{align}
    \mathcal{L}_{u}^A=\frac{1}{B_u} \sum_{i=1}^{B_u} \mathbbm{1} (\max (p_i^F) \geq \tau) \mathcal{H}(\hat{y}_i^F, A(\mathcal{T}_{strong}(u_i))). \label{eqa-unsupa}
\end{align}

\noindent\textbf{Complete Training Objective.}
To summarize, with the supervised losses from labeled data and the unsupervised losses from unlabeled data, the complete objective function is presented as
\begin{align}
    \mathcal{L} = (\mathcal{L}_{s}^F+\mathcal{L}_{s}^A) + \lambda (\mathcal{L}_{u}^F+\mathcal{L}_{u}^A), \label{eqa-totalloss}
\end{align}
where $\lambda$ denotes the balancing coefficient for the unsupervised losses.
Note that the size of the auxiliary network $A(\cdot)$ is designed to be much smaller than the backbone $F(\cdot)$; therefore, the introduction of $A(\cdot)$ has little effect on training efficiency.

\subsection{Implementation} \label{sec-method-implementation}

In this part, we describe the implementation of CMPL.

\noindent \textbf{Auxiliary Network.} As stated in \cref{sec-method-cmpl}, the auxiliary network should have a different capacity from the primary network to provide complementary representations. In practice, we obtain the auxiliary network by adjusting the depth and width of the primary network. For instance, if the primary network is 3D-ResNet50, the auxiliary network could be 3D-ResNet18 or 3D-ResNet50$\times 1/4$ where the channels are reduced four-fold. Comprehensive ablation studies on the depth and width of the auxiliary network are included in \cref{tab-ablation-arch} of the experiments section. Unless otherwise specified, we adopt 3D-ResNet50 and 3D-ResNet50$\times 1/4$ as the primary backbone and auxiliary network, respectively.

\noindent \textbf{Temporal Data Augmentations.}
For spatial data augmentation, we strictly follow the augmentations of \cite{nonlocal, slowfast} for training as $\mathcal{T}_{standard}(\cdot)$ on labeled data. For unlabeled data,
center cropping with a patch size of 224 $\times$ 224 is employed as the weak augmentation $\mathcal{T}_{weak}(\cdot)$.
Following FixMatch \cite{sohn2020fixmatch}, RandomAugmentation \cite{uda, sohn2020fixmatch} together with Cutout \cite{devries2017cutout} are used as the strong augmentations $\mathcal{T}_{strong}(\cdot)$. 
Note that all the spatial transformations of the above augmentations are temporally consistent across all frames in the same video clip.

Besides spatial augmentations, commonly used temporal data augmentations, namely various \emph{temporal locations} and \emph{frame rates}, are compatible with our CMPL. Previous work~\cite{feichtenhofer2021large} suggests that visual content is often temporally persistent throughout the time span of a video, such that different clips of a given video share similar representations. Accordingly, we adjust the temporal location of the input clip and control the augmentation strength via an adjustable \textbf{time offset $t_s$}. In our CMPL framework, this is done by feeding the primary backbone $F(\cdot)$ and the auxiliary network $A(\cdot)$ clips from different temporal locations of the same video while still requiring them to supervise each other. Meanwhile, we also follow \cite{vthcl,tcl} in regarding different frame rates as a form of temporal augmentation. This is also illustrated in \cref{fig-framework}, where a faster frame rate results in more frames given that all clips share the same temporal duration.

\section{Experiments}\label{sec-experiment}

\begin{table*}[t]
  \caption{
    \textbf{Comparison with state-of-the-art methods on UCF-101 and Kinetics-400.} Note that R18 and R50 denote the backbone networks and their depths. We report top-1 accuracy as the evaluation metric. “Input” shows the input data format used during training, where
    "V" is the raw rgb video,  "F" is optical flow and "G" is the temporal gradient.}
 \label{tab-system-level}
 \vspace{-5pt}
  \centering
\setlength{\tabcolsep}{15pt}
 \begin{tabular}{llcccccc}
 \toprule
  \multirow{2.5}{*}{Method} & \multirow{2.5}{*}{Backbone} & \multirow{2.5}{*}{Input} & \multirow{2.5}{*}{\#Frames}    & \multicolumn{2}{c}{UCF-101} & \multicolumn{2}{c}{Kinetics-400} \\ \cmidrule{5-8} 
                            &    &   &                          & 1\%          & 10\%         & 1\%            & 10\%            \\ \midrule
Supervised & 3D-ResNet50 &V & 8  & 6.5 & 32.4 & 4.4 & 36.2 \\ \midrule
\multicolumn{7}{l}{\textit{Image-based Semi-supervised Methods}} \\
S4L~\cite{zhai2019s4l} & 3D-ResNet18  & V  & 16   &              -     &   29.1   & -               &     -      \\
FixMatch~\cite{sohn2020fixmatch} &  3D-ResNet50 & V & 8  & 14.8 & 49.8 & 8.6  &  46.9         \\ 
FixMatch~\cite{sohn2020fixmatch} &  SlowFast-R50 &V& 8  & 16.1 & 55.1 & 10.1  & 49.4         \\ 
\midrule
\multicolumn{7}{l}{\textit{Video-based Semi-supervised Methods}} \\
VideoSSL~\cite{jing2021videossl} &  3D-ResNet18 &V & 16 &                -     &   42.0   &       -         &       -    \\ 
TCL~\cite{tcl} & TSM-R18 &R & 8  &              -     &   -      &  8.5    & -       \\ 
MvPL~\cite{xiong2021multiview} &    3D-ResNet50 & V+F+G  & 8    &    22.8 &   \textbf{80.5}    &  17.0    & 58.2                 \\ \midrule
Ours &       3D-ResNet18 &V & 8    & 23.8 &   67.6    & 16.5     & 53.7                  \\
Ours &       3D-ResNet50 &V  & 8    & \textbf{25.1} &   79.1   & \textbf{17.6}   & \textbf{58.4} \\ \midrule

\end{tabular}
\vspace{-5pt}
\end{table*}

We evaluate the proposed CMPL on two commonly used datasets, namely Kinetics-400 \cite{kinetics} and UCF-101 \cite{soomro2012ucf101},
using two standard settings for semi-supervised action recognition (\ie~1\% and 10\% labeled data).
Detailed ablation studies on the design choices of CMPL are also included.
In addition, some empirical analysis is presented to validate our motivations for CMPL.
We note that all the experiments are conducted on a single modality (\emph{i.e.}, RGB) and evaluated on the corresponding validation set unless otherwise stated.

\subsection{Settings} \label{sec-implementation}

\noindent{\textbf{Dataset.}} Kinetics-400 \cite{kinetics} consists of 240K and 20K videos respectively for training and validation, 
covering 400 categories.
Each video lasts about 10 seconds.
UCF-101 \cite{soomro2012ucf101} contains 101 action classes with roughly 9.5K training videos and 4K validation videos.
To obtain 1\% or 10\% labeled data for the two evaluation settings,
for Kinetics-400 we form two balanced labeled subsets by sampling 6 or 60 videos per class. 
As for UCF101, we sample 1 or 10 videos per class as labeled data.
To reduce the randomness brought by the sampling process, the accuracy averaged over \textbf{three different sampled subsets} is reported.
To obtain the input frames for each video,
we first extract 64 consecutive frames as a raw clip,
then sample sparsely with a stride $s$ to obtain $64/s$ frames, denoted as $64/s \times s$. 
Unless stated otherwise, the input frames for the primary and auxiliary branches are set to $8\times8$ and $16\times4$, respectively.

\noindent{\textbf{Training.}}
For Kinetics-400, we train the model using 16 GPUs, with the SGD optimizer, momentum 0.9, and weight decay 0.0001.
The batch size for labeled data and unlabeled data are 2 and 10 on each GPU, resulting in a mini-batch size of 192 in total. 
The base learning rate is set to 0.1 and decayed according to the cosine scheduler. A total of 200 training epochs is used.
In addition, $\lambda$ is set to 5 and 2 for the 1\% and 10\% labeled data settings, and $\tau$ is set to 0.9.
The training setting of UCF-101 is the same as that for Kinetics-400 except that the weight decay is set to $0.001$.

\noindent{\textbf{Inference.}}
Following the test protocol in \cite{nonlocal, slowfast, tpn}, we uniformly sample ten clips over the whole video and average the softmax probabilities of all clips as the final prediction.
The shorter spatial side of the input video is scaled to 256 pixels, and three crops of size 256 $\times$ 256 are extracted to obtain more spatial information. 
Although the primary and auxiliary network are jointly optimized in training, we use only the primary model for inference, leading to no additional inference cost.

\noindent{\textbf{Baselines.}}
First, we have a supervised baseline with the same architecture as our approach and the same set of labeled samples for training, but without any unlabeled data.
Secondly, we extend the state-of-the-art semi-supervised learning approaches in the image domain \cite{zhai2019s4l, sohn2020fixmatch} to the video domain for comparison. For Fixmatch \cite{sohn2020fixmatch}, we adopt the same experimental settings as our approach.
Finally, we also include the state-of-the-art video-based semi-supervised learning methods \cite{jing2021videossl, tcl, xiong2021multiview}, whose performances from their original papers are directly reported.

\subsection{Main Results} \label{sec-experiment-mainresults}
Quantitative results are presented in \cref{tab-system-level}. Compared to previous methods that use a single modality of RGB frames (all baselines except for MvPL \cite{xiong2021multiview}), our CMPL leads by a clear margin even when equipped with a shallower network, \emph{i.e.}, 3D-ResNet18 as the backbone. Further improvements are obtained by increasing the capacity of the backbone network \emph{i.e.}, from 3D-ResNet18 to 3D-ResNet50. 
 
While CMPL outperforms FixMatch when both of these methods adopt 3D-ResNet50 as the backbone,
the improvement of CMPL may come from the inclusion of the lightweight auxiliary network.
To investigate this,
we change the backbone of FixMatch from 3D-ResNet50 to SlowFast-R50, which is a two-branch network that also includes a lightweight component for feature aggregation.
As shown in \cref{tab-system-level},
CMPL still outperforms FixMatch with a larger backbone,
suggesting it is the auxiliary supervision that matters.   

Meanwhile, the concurrent work MvPL~\cite{xiong2021multiview} achieves performance competitive to ours. 
However, for these results it uses 3 modalities, compared to only the RGB modality for CMPL.

\subsection{Ablation Study} \label{sec-experiments-ablation}

In this part, we make comprehensive analysis of the proposed CMPL via ablation studies.
We first analyze the key components of our approach in \cref{sec-expriments-cmplcomponents}, followed by the experimental results of different instantiations of losses and hyperparameters in \cref{sec-expriments-hyperparameters}.
Note that all ablation studies are performed on \textbf{1\% labeled data of Kinetics-400 with 50 training epochs.}
The 3D-ResNet50 backbone and the sparse sampling strategy (8 $\times$ 8) are adopted for the primary backbone unless specified otherwise.

\subsubsection{CMPL Components} \label{sec-expriments-cmplcomponents}

\begin{table}[t]
    \caption{\textbf{Ablation studies on CMPL components.} We gradually add the auxiliary network (Auxilary), different visual tempos (Frame Rates) and different temporal locations (Temporal Loc)
    to the baseline model (FixMatch). All studies are trained with \textbf{50 epochs on 1\% labeled Kinetics-400}. Unless specified, the results of following tables are all with the same training configuration.}
    \label{tab-ablation-components}
    \vspace{-5pt}
    \centering
        \begin{tabular}{ccccc}
        \toprule
        Auxilary   &  Frame Rates & Temporal Loc     & Top-1   \\ \midrule
        \multicolumn{3}{c}{\textit{FixMatch}}        & 6.78    \\ 
                  & \ding{51}       &               & 7.12   \\ 
                  & \ding{51}       & \ding{51}    &  7.68        \\\midrule
        \ding{51}  &                 &               & 12.04   \\ 
        \ding{51}  & \ding{51}       &               & 12.90   \\
        \ding{51}  & \ding{51}       & \ding{51}     & \textbf{13.71}   \\ \bottomrule
        \end{tabular}
    \vspace{-10pt}
\end{table}

\noindent{\textbf{Ablations on CMPL components.}}
While the auxiliary network provides pseudo-label estimates that complement those generated by the primary backbone,
we adopt two advanced temporal augmentations to further enhance divergence between the representations of these two networks, to encourage greater complementarity. 
As shown in \cref{tab-ablation-components},
starting from the baseline approach FixMatch which achieves 6.78\% Top-1 accuracy, introducing the auxiliary supervision improves the accuracy to 12.04\%, demonstrating the effectiveness of the cross-model strategy. On top of this, the temporal augmentations further enhance performance to 13.71\%, leading to a stronger result.

\begin{table}
        \caption{\textbf{Architectures of the auxiliary network.} Different architectures are adopted as the auxiliary network to study the effects of network architectures. Specifically, when the auxiliary pathway adopts the primary's architecture, CMPL degenerates to FixMatch~\cite{sohn2020fixmatch}, providing a strong baseline for our comparison. }

        \label{tab-ablation-arch}

    \centering
        \begin{tabular}{lcccccc}
        \toprule
        Auxiliary Network & \#Frames    & \#FLOPs   & Top-1   \\\midrule
        Primary               & 8             &  146G     &  6.78            \\ 
        Primary               & 16            &  168G     &   7.12  
        \\ 
        3D-ResNet50      & 16                & 187G & 9.97          \\\midrule
        \multicolumn{4}{c}{\textit{network depth}}         \\ 

        3D-ResNet18      & 16                & 132G& 10.63  \\
        3D-ResNet10      & 16                & 108G & 10.88    \\ \midrule
        \multicolumn{4}{c}{\textit{network width}}         \\ 
        3D-ResNet50$\times 1/2$        & 16            &  112G     & 11.57               \\
        3D-ResNet50$\times 1/4$        & 16             &  94G   & \textbf{12.90}              \\
        3D-ResNet50$\times 1/8$        & 16             &  87G    & 12.08             \\ 
        \bottomrule

        \end{tabular}
\end{table}

\noindent{\textbf{Auxiliary network architecture.}}
To study the effect of auxiliary network architecture,
we start by using the same architecture for both the backbone and the auxiliary network, which leads to two different versions, one with a shared set of parameters (equivalent to FixMatch) and another with two different sets of parameters, corresponding to the 2nd and 3rd lines of \cref{tab-ablation-arch}.

Subsequently, we also shrink the depth and width of 3D-ResNet50 to produce other auxiliary network variants.
\cref{tab-ablation-arch} presents the performance with different auxiliary networks. It can be seen that even with the backbone and auxiliary network being identical but with different weights,
learning in the cross-model manner still leads to performance improvements. 
Moreover,
substantial gains are obtained no matter which factor (\emph{e.g.}, depth or width) is shrunk,
indicating that a relatively small network as the auxiliary network can provide complementary information for the backbone.
In addition, the cross-model learning strategy is the key to the enhanced performance, rather than the newly introduced model capacity, since consistent gains are observed as the auxiliary network varies.

\begin{table}[t]
    \setlength{\tabcolsep}{10pt}
    \caption{\textbf{Effects of input \#Frames.} Different number of frames are adopted as the auxiliary input to evaluate the effect of frame rates in CMPL.}
    \label{tab-ablation-frames}
    \vspace{-5pt}
    \centering
    \begin{tabular}{x{32}x{32}x{32}x{32}}
    \toprule
     $T$    &  8 $\times$ 8 & 16 $\times$ 4 & 32 $\times$ 2 \\\midrule
     Top-1    & 12.04   & 12.90 & 12.94 \\ \bottomrule
    \end{tabular}
    \vspace{-10pt}
\end{table}

\noindent{\textbf{Frame rate for visual tempo.}}
To study how the frame rates between the primary backbone and auxiliary network affect the performance, we fix the sampling rate (8 $\times$ 8) for the backbone, and use different sampling rates for the auxiliary network, namely 8 $\times$ 8, 16 $\times$ 4, and 32 $\times$ 2.

As shown in \cref{tab-ablation-frames}, a larger sampling rate leads to a better performance.
As stated in previous work~\cite{tpn, slowfast}, a larger difference between two sampling rates usually leads to a more different representations of the same videos at the input level.
This naturally constructs two relevant and complementary views, further boosting our cross-model learning.

\noindent{\textbf{Time offset for temporal location.}}
Besides frame rate, we also explore the effect of the time offset for temporal location, which is controlled by the hyper-parameter $t_s$ as shown in \cref{sec-method-implementation}. 
\cref{tab-ablation-timespan} reports the performance for increasing $t_s$.
It is seen that a larger $t_s$ positively affects the final performance. 
It should be noted that the trimmed videos of Kinetics last about 10s so that $ts$ is set to at most 10. 

\begin{table}
    \caption{\textbf{Effects of time offset $t_s$.} The time offset determines the temporal distance between the clips processed by the primary and auxiliary networks.}
    \label{tab-ablation-timespan}
    \vspace{-5pt}
    \centering
    \begin{tabular}{x{26}x{26}x{26}x{26}x{26}}
    \toprule
     $t_s$    & 1 & 2 & 5 & 10 \\ \midrule
     Top-1    & 12.90 & 12.64   & 13.44  & 13.71 \\ \bottomrule
    \end{tabular}
    \vspace{0pt}
\end{table}

\begin{table}[t]
    \caption{\textbf{Effects of the losses for the auxiliary network.}}
    \label{tab-ablation-losses}
    \vspace{-5pt}
    \centering
    \begin{tabular}{x{36}x{36}x{36}x{36}}
    \toprule
      $\mathcal{L}_{s}^A$       & $\mathcal{L}_{u}^A$ & Top-1 & Top-5   \\ \midrule
      
        \multicolumn{2}{c}{\textit{FixMatch}}  & 6.78 &  17.92  \\ 
            \ding{51}                   &            & 10.57    & 25.53 \\
                                      & \ding{51}  & 10.32   & 24.46\\
    \ding{51}                       & \ding{51}  & 12.90  &  28.19  \\ \bottomrule
    \end{tabular}
    \vspace{-5pt}
\end{table}

\subsubsection{Choices of losses and hyperparameters} \label{sec-expriments-hyperparameters}

\noindent{\textbf{Losses for auxiliary branch.}}
There is a supervised and an unsupervised loss for both the primary and the auxiliary networks. Here, we examine the importance of the supervised and unsupervised loss of the auxiliary network, while the two losses of the primary network are kept.

\cref{tab-ablation-losses} presents the results. Including the auxiliary network always improves performance, no matter where the labels come from (\emph{e.g.} pseudo-labels or ground-truth labels). This demonstrates that the auxiliary network brings representations complementary to those of the primary network. Including both types of labels leads to the best results.

\begin{figure}[t]
    \centering
    \subfloat[\textbf{Threshold.} \label{tab-ablation-threshold}]{
     \includegraphics[width=0.3\linewidth]{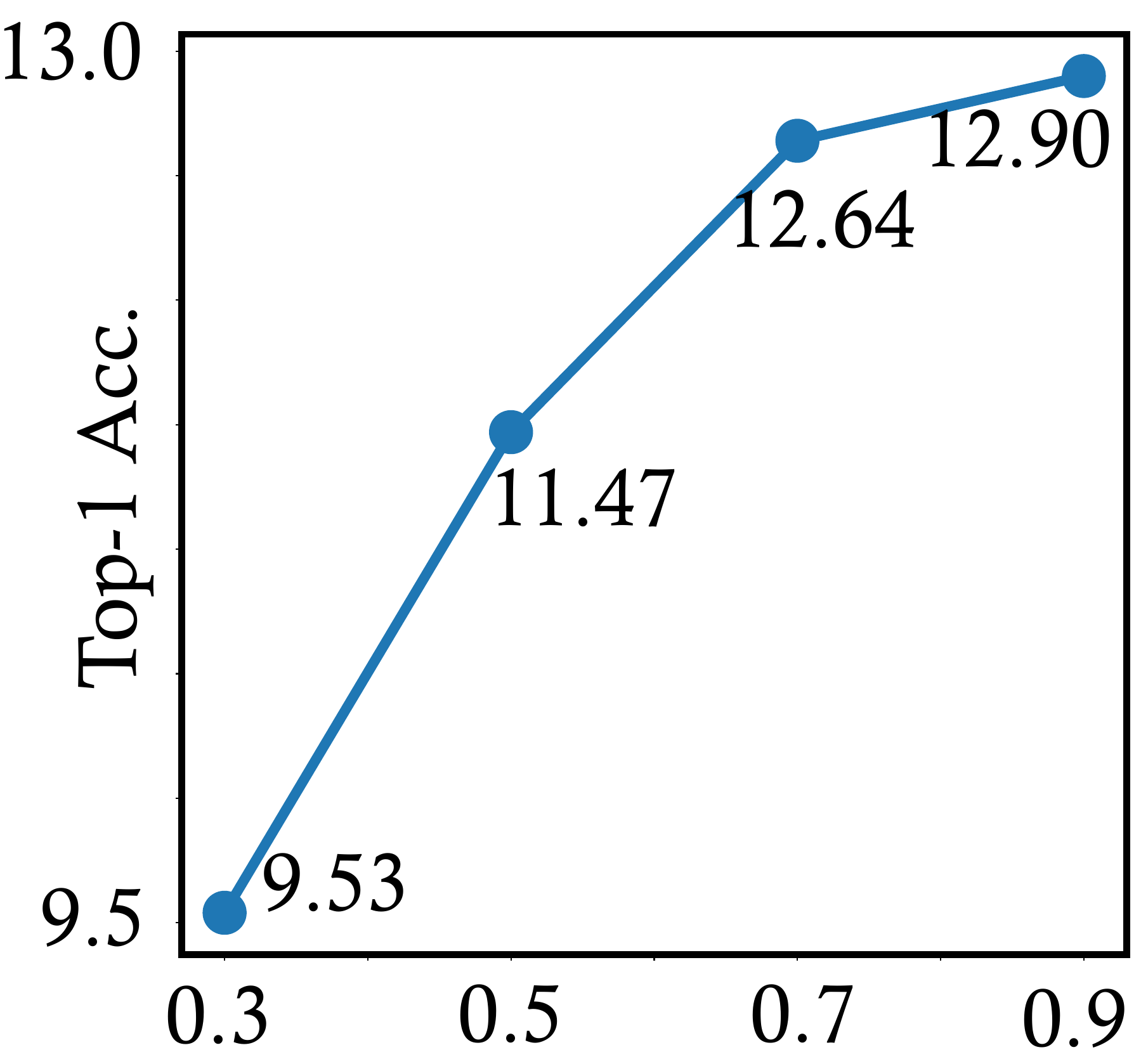}
    }
        \subfloat[\textbf{Batch ratio.} \label{tab-ablation-batchratio}]{
     \includegraphics[width=0.3\linewidth]{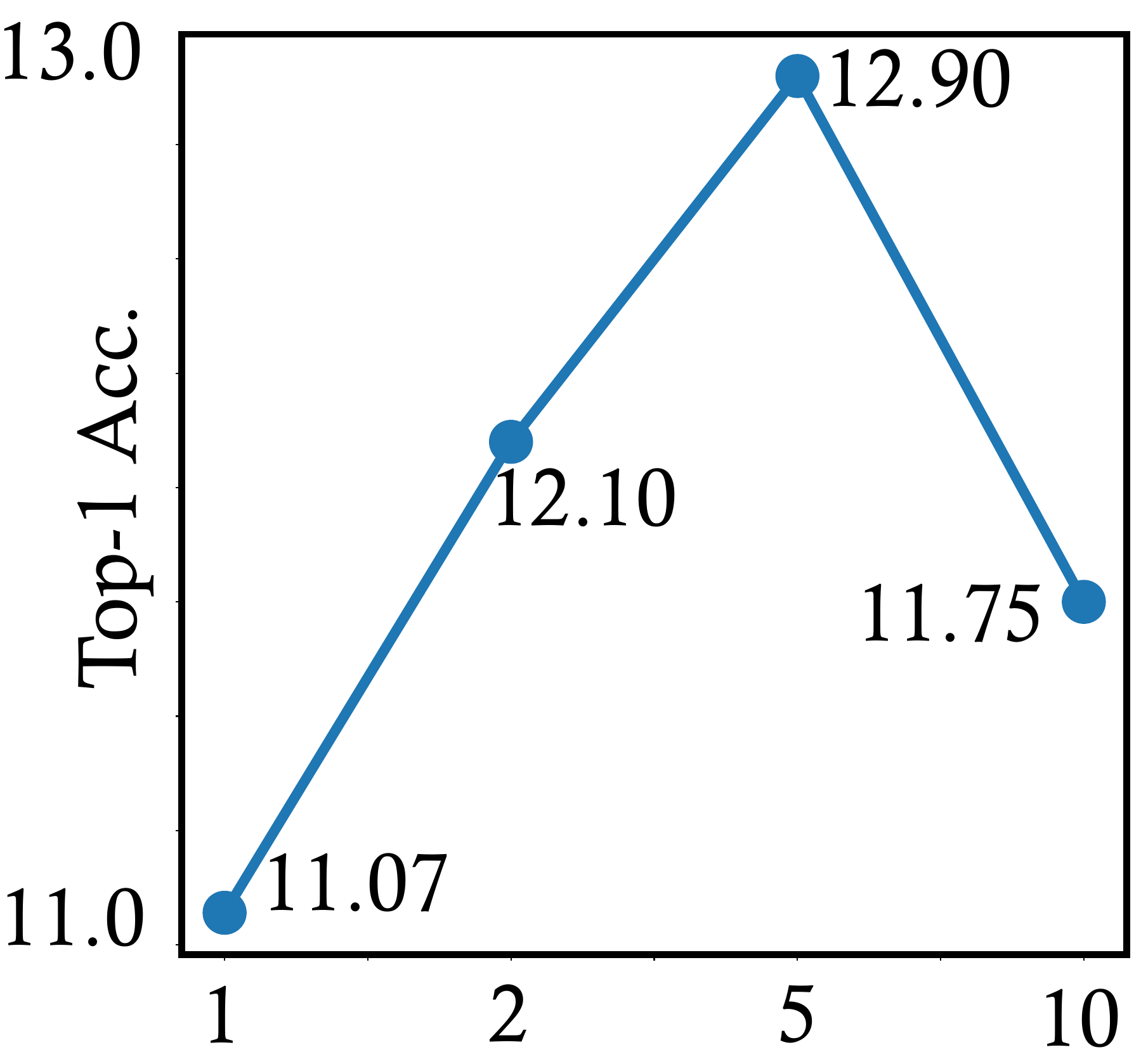}
    }
        \subfloat[\textbf{Loss weight.} \label{tab-ablation-lossweight}]{
     \includegraphics[width=0.3\linewidth]{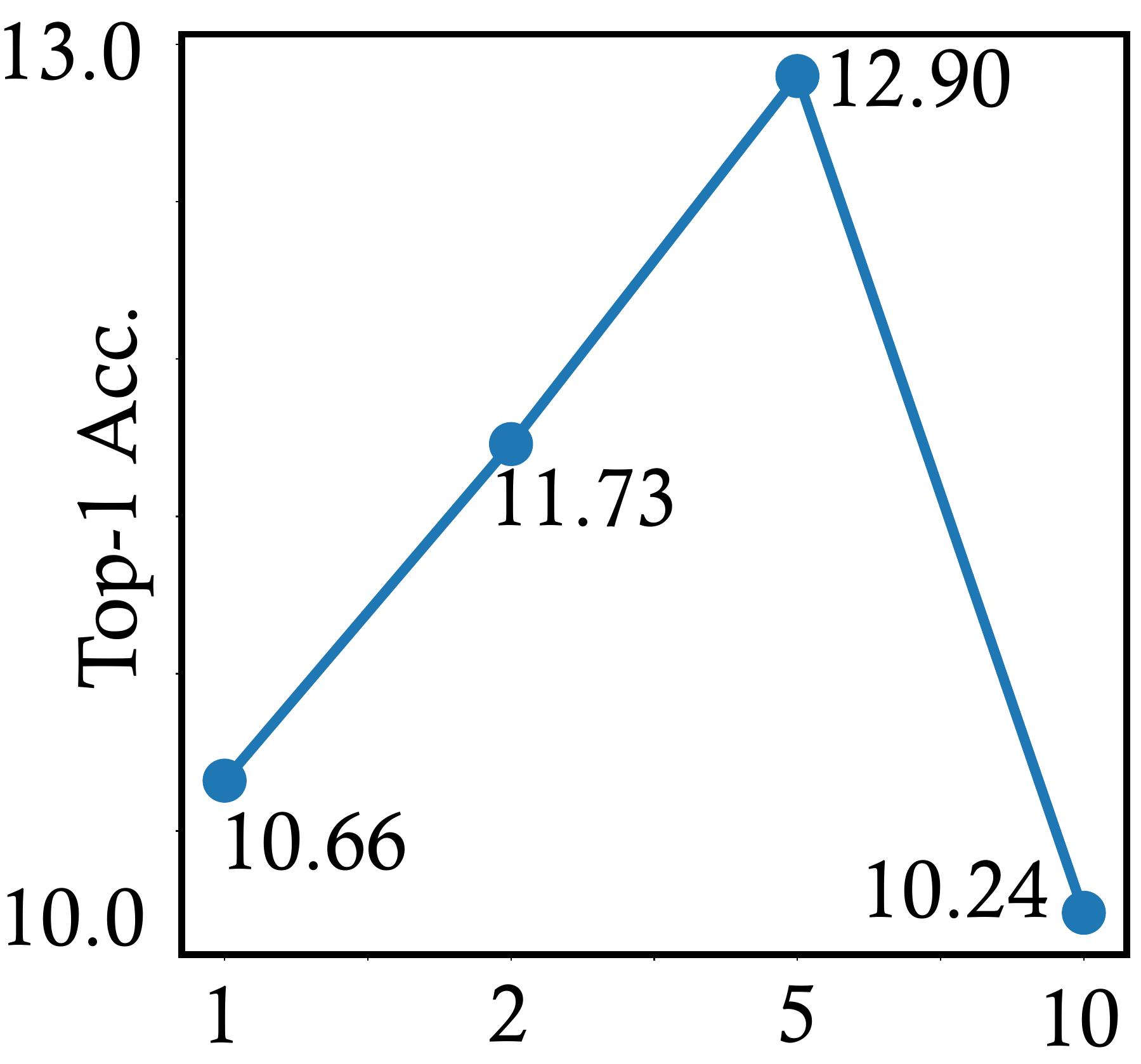}
    }
    \caption{\textbf{Effects of hyperparameters.} Results of varying threshold, the ratio of unlabeled data to labeled data ($B_u/B_l$), as well as the loss weight ($\mu$), are included to comprehensively study the effects of the hyperparameters.}
\end{figure}

\noindent{\textbf{Hyperparameters.}}
\cref{tab-ablation-threshold} presents the effects of different thresholds,
where using 0.9 as the threshold obtains the best performance.
It is concluded that a higher threshold leads to better performance, \emph{i.e.}, the quality of the pseudo labels matters.
Besides, we also consider the ratio of unlabeled data to labeled data ($B_u/B_l$). We fix the batch size of labeled data to be 2 and set the ratio to be $\{1, 2, 5, 10\}$ to analyze its effect. As shown in \cref{tab-ablation-batchratio}, $B_u/B_l$ set to be 5 can achieve the best trade-off between performance and computation cost.
We also analyze the effect of the loss weight $\lambda$ and observe that it affects the performance as accuracy drops by 2\% when setting $\lambda$ to 1 or 10. \cref{tab-ablation-lossweight} illustrates that the optimal value of the loss weight is 5, at which the supervised loss and unsupervised loss is properly balanced.

\begin{figure}[t]
    \centering
    \includegraphics[width=0.95\linewidth]{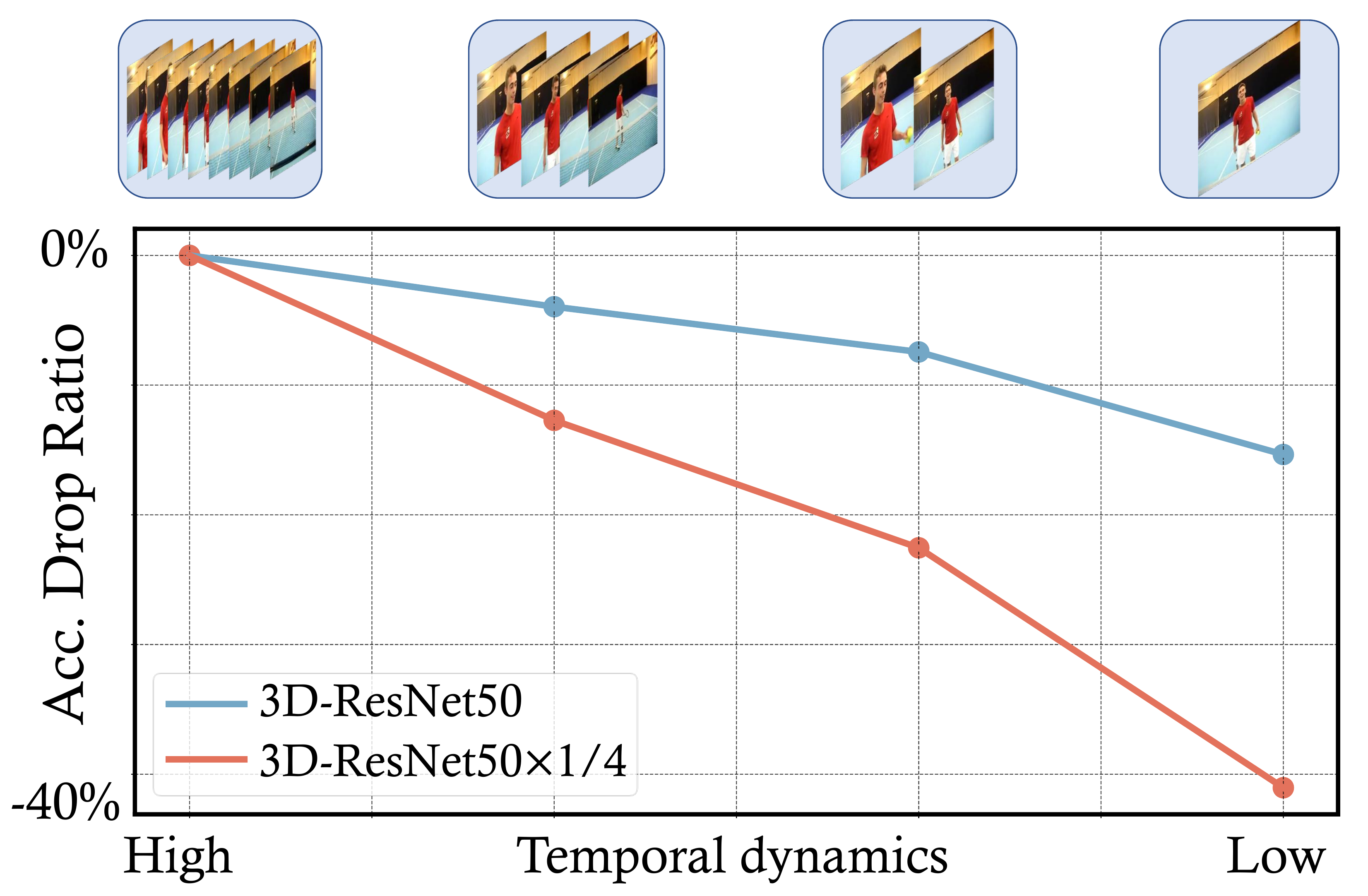}
    \vspace{-5pt}
    \caption{\textbf{Accuracy drop ratio with different frames per clip.} The top half of this figure presents the re-sampled frames with different strides. As the stride increases, the number of re-sampled frames decreases, weakening the temporal dynamics of the given clip. We evaluate the performance drop ratio of the re-sampled frames to study the representations learned by the two different networks.}
    \label{fig-dynamics}
    \vspace{-15pt}
\end{figure}

\subsection{Empirical Analysis} \label{sec-experiment-empirical}
To validate our motivations for CMPL, some empirical analysis is conducted on Kinetics-400 with 1\% labeled data. 
Unless specified otherwise, we use 3D-ResNet50 as the backbone, which takes 8 frames as input.
As for the auxiliary network, we use 3D-ResNet50$\times$1/4, which takes 16 frames as input.

\noindent{\textbf{Representation learned by different networks.}}\label{sec-experiment-empirical-learned}
\cref{fig-observation} indicates that networks of different scale learn complementary semantics with limited annotated data.
To study the mechanism by which these two networks complement each other,
we first train the primary and auxiliary networks separately with 8 $\times$ 8 frames as the input. They are only trained with 1\%  labeled  samples of Kinetics-400. We then re-sample the 8 frames with different stride $\{1, 2, 4, 8\}$ in order to decrease the temporal dynamics of a given video clip. Notably, no matter how many frames the sampled clip contains, they are always extended to a 8-frame clip for testing.

\cref{fig-dynamics} displays accuracy curves of the two networks with inputs of varying temporal dynamics. It can be seen that the auxiliary network is more sensitive to temporal dynamics. This suggests that the tiny model using a lower channel capacity can better capture fast motion without building a detailed spatial representation, providing complementary temporal semantics in relation to the spatial patterns learned by the backbone network.

\definecolor{myOrange}{RGB}{221, 132, 82}
\definecolor{myGreen}{RGB}{85, 168, 104}
\definecolor{myBlue}{RGB}{76, 114, 176}
\definecolor{myRed}{RGB}{227, 114, 92}
\definecolor{myBlue2}{RGB}{115, 167, 198}

\noindent{\textbf{Effects of the auxiliary pseudo-labels.}}
A key difference between CMPL and FixMatch~\cite{sohn2020fixmatch} is the source of the pseudo-labels. The primary backbone takes the pseudo-labels from the auxiliary network while FixMatch only uses its own high-confidence predictions. To study the quality of the pseudo-labels provided by the auxiliary network, we compare three sources of pseudo-labels, namely the backbone network, the auxiliary network trained in CMPL, and the 3D-ResNet50 trained in FixMatch, which has the same architecture as the backbone network. 
We first select the samples with pseudo-labels assigned by the auxiliary network to build a subset. And then the accuracy of these three pseudo-label sources are tested on this subset. Accuracy is recorded over the whole training process at an interval of 10 epochs.

\begin{figure}[t]
    \centering
    \includegraphics[width=0.95\linewidth]{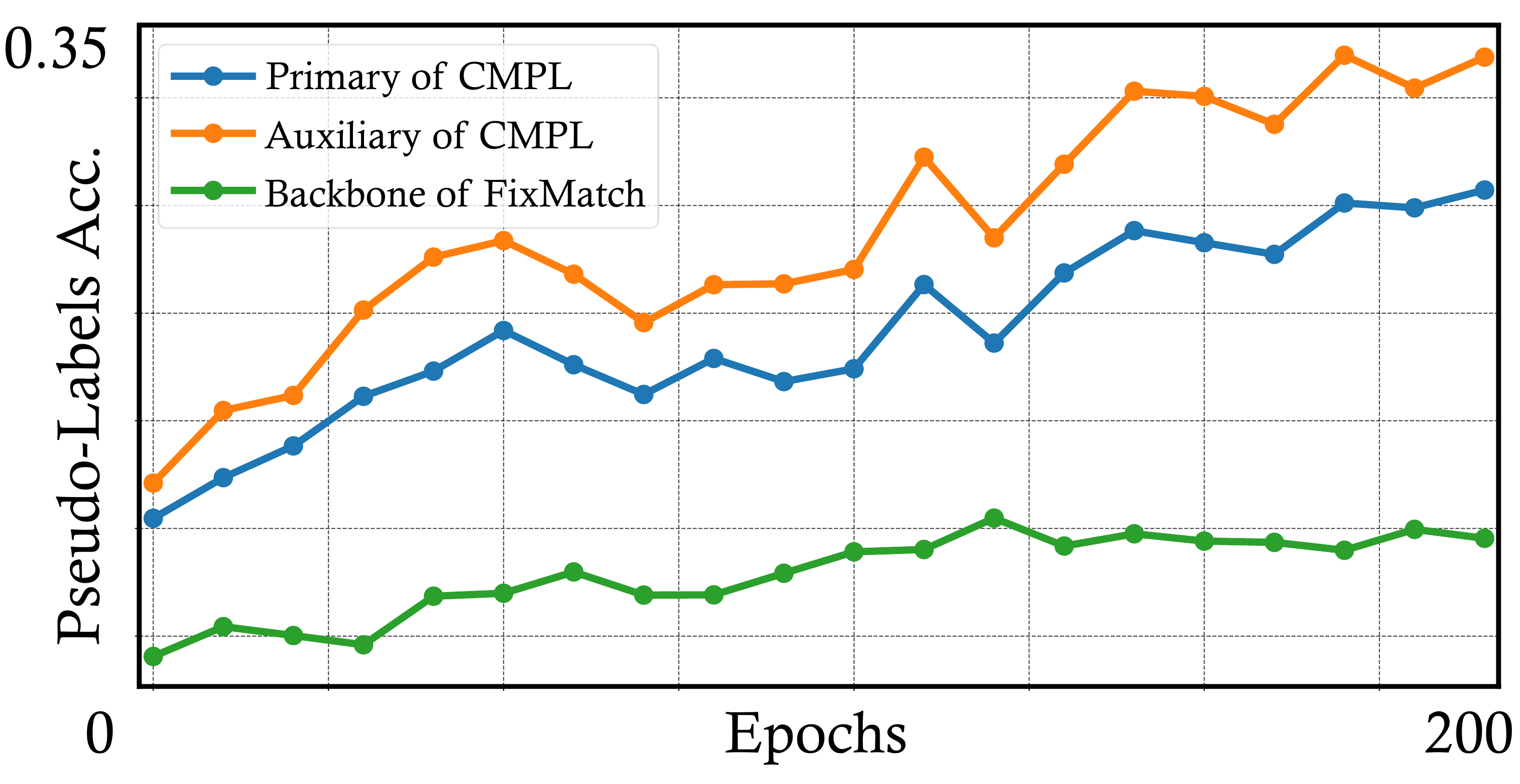}
    \vspace{-15pt}
    \caption{{Training accuracy curves} of the primary branch of CMPL (\textbf{\textcolor{myBlue}{blue}}), the backbone of FixMatch (\textbf{\textcolor{myGreen}{green}}), and the auxilary of CMPL (\textbf{\textcolor{myOrange}{orange}}). Training accuracy is evaluated on the samples with pseudo-labels assigned by the auxiliary network.
    Please refer to \cref{sec-experiment-empirical} for more details.}
    \label{fig-trainingcurve} 
    \vspace{-10pt}
\end{figure}

\cref{fig-trainingcurve} shows the accuracy curves. Obviously, for the samples originally labeled by the auxiliary network with high confidence,
the auxiliary network consistently obtains better pseudo-labels (\textbf{\textcolor{myOrange}{orange curve}}) than the backbone network (\textbf{\textcolor{myBlue}{blue curve}}) throughout the training process. 
Moreover,
compared to the same architecture learned with FixMatch (\textbf{\textcolor{myGreen}{green curve}}), the backbone network gradually produces better estimates for these selected samples.
Such results demonstrate 1) cross-model pseudo-labeling using a tiny auxiliary network can enhance the backbone network; 2) the backbone network indeed learns from the auxiliary network through the proposed cross-model strategy; 3) although the low-capacity auxiliary network tends to perform worse than the high-capacity backbone network, it still provides knowledge to guide the latter.

\begin{figure}
    \centering
	\includegraphics[width=0.95\linewidth]{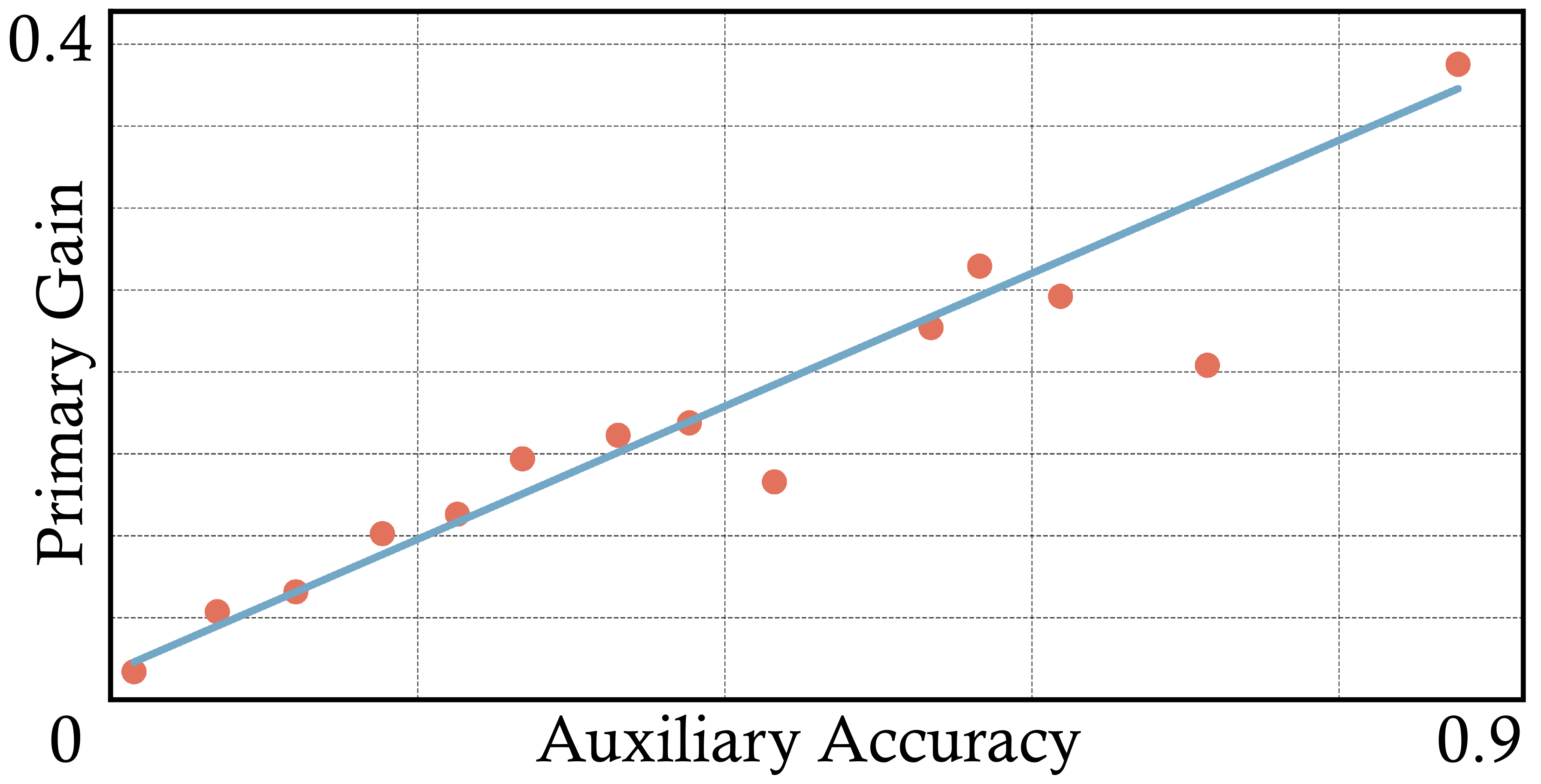}
	\vspace{-10pt}
    \caption{\textbf{Primary gain \emph{vs.}~Auxiliary accuracy.} Each \textbf{\textcolor{myRed}{red}} point denotes the mean accuracy gain of the primary backbone within a bin of auxiliary network accuracy. 
    The \textbf{\textcolor{myBlue2}{blue}} line plots a least squares approximation.}
     \label{fig-correlation}
    \vspace{-15pt}
\end{figure}

\noindent \textbf{Primary gain \emph{vs.} Auxiliary accuracy.}
The key idea of CMPL is to adopt an auxiliary network to provide complementary information that promotes better representation.
Therefore, we examine the per-class gain (\emph{primary gain}) of the backbone network on the Kinetics-400 validation dataset when adopting our CMPL instead of FixMatch. We also include the per-class performance of the auxiliary network (\emph{auxiliary accuracy}) trained in a supervised manner as a reference.
For better visualization, we discretize auxiliary accuracy into bins with an interval of 0.05 and then calculate the mean of the primary gain (y-axis) in each bin.
As shown in \cref{fig-correlation}, the performance gain of the backbone network is positively correlated with that of the auxiliary network on the corresponding classes.
This study supports our motivation that the auxiliary network complements the backbone network, particularly on action classes recognized well by the former. 

\vspace{-5pt}
\section{Conclusion}\label{sec-conclusion}
\vspace{-5pt}
We present a Cross-Model Pseudo-Labeling framework for semi-supervised action recognition. It pairs a lightweight auxiliary network with the primary backbone such that they can learn from each other via pseudo-labeling.
Over multiple datasets and low data regimes, our approach outperforms its supervised counterpart with limited labeled data by a substantial margin and surpasses several previous methods using only RGB frames.
Moreover, we show that the primary network can be complemented by an auxiliary network even with the same input, 
demonstrating a new alternative to the previous standard of changing the input data in semi-supervised learning.
While CMPL obtains improved performance for semi-supervised action recognition, its results with 10\% annotated data are not as good as supervised training with 100\% labeled data.
This leaves room for further investigation into improving CMPL with more advanced techniques.

\noindent\textbf{Acknowledgement} We thank Nanxuan Zhao for discussion and comments about this work.

{\small
\bibliographystyle{ieee_fullname}
\bibliography{ref}
}

\newpage 

\appendix
\newcommand{\AppendixPrefix}{A}
\renewcommand{\thefigure}{\AppendixPrefix\arabic{figure}}
\setcounter{figure}{0}
\renewcommand{\thetable}{\AppendixPrefix\arabic{table}} 
\setcounter{table}{0}
\renewcommand{\theequation}{\AppendixPrefix\arabic{equation}} 
\setcounter{equation}{0}

\newcommand{\blocks}[3]{\multirow{3}{*}{\(\left[\begin{array}{c}\text{3$\times$1$\times$1, #2}\\[-.1em] \text{1$\times$3$\times$3, #2}\\[-.1em] \text{1$\times$1$\times$1, #1}\end{array}\right]\)$\times$#3}
}

\section*{Appendix}

\section{Pseudo-Labeling Schemes}\label{sec-method-instantiation}
How to generate pseudo-labels for unlabeled data based on the model outputs is also an important question for CMPL.
Many pseudo-labeling strategies become possible with the introduction of the auxiliary network.
Besides the one described in our method, we list some other options:
    1) \textit{Self-First}: Each network first checks whether its own prediction is confident enough, and if it is not, then the label is obtained from its sibling.
    %
    2) \textit{Opposite-First}: Each network instead prioritizes its companion over itself.
    %
    %
    3) \textit{Maximum}: The most confident prediction from the two networks is taken as the pseudo-label.
    %
    4) \textit{Average}: The predictions from the two networks are averaged before deriving the pseudo-label.

Let the pseudo-label confidence produced by $F$ and $A$ be $l_F$ and $l_A$.
The pseudo-label confidences for a video $u_i$ are thus $l_F(u_i) = p_i^A, l_A(u_i) = p_i^F$. 
And now the corresponding mathematical formulations of different pseudo-labeling schemes are presented as following, where $u_i$ is removed for clarity.
\begin{enumerate}

\item \textit{Self-First}:
\begin{align}
    l_F = \mathbbm{1} (\max (p^F) \geq \tau)p^F + (1-\mathbbm{1} (\max (p^F) \geq \tau))p^A, \nonumber \\
    l_A = \mathbbm{1} (\max (p^A) \geq \tau)p^A + (1-\mathbbm{1} (\max (p^A) \geq \tau))p^F. \nonumber
\end{align}

\item \textit{Opposite-First}:
\begin{align}
    l_F = \mathbbm{1} (\max (p^A) \geq \tau)p^A + (1-\mathbbm{1} (\max (p^A) \geq \tau))p^F, \nonumber \\
    l_A = \mathbbm{1} (\max (p^F) \geq \tau)p^F + (1-\mathbbm{1} (\max (p^F) \geq \tau))p^A. \nonumber
\end{align}

\item \textit{Maximum}:
\begin{align}
    l_F = l_A = \mathbbm{1} (\max (p^F) \geq \max (p^A)) p^F + \nonumber \\ (1-\mathbbm{1} (\max (p^F) \geq \max (p^A))) p^A. \nonumber
\end{align}

\item \textit{Average}:
\begin{align}
    l_F = l_A = \frac{p^F+p^A}{2}. \nonumber
\end{align}
\end{enumerate}

  \begin{table*}[t]
    \caption{
        \textbf{Comparison with other state-of-the-art self-supervised learning methods on UCF-101.} 
        We use UCF-101 as the labeled data and Kinetics-400 as the unlabeled data.
        The other self-supervised methods are pretrained on Kinetics-400 and fine-tuned on UCF-101.
        Our model are trained from scratch.
    }
    \label{table:sota}
    \vspace{-5pt}
    \centering
    \setlength{\tabcolsep}{12pt}{
    \begin{tabular}{llcc}
      \toprule
      Method                             & Architecture  & \#Frames  & UCF-101 \cite{soomro2012ucf101}  \\
      \midrule
      Random Init          &  3D-ResNet50  & 8         & 61.1  \\
      ImageNet Init         &  3D-ResNet50  & 8         & 86.2  \\
      \midrule
      MotionPred \cite{motionpred}       &  C3D\cite{c3d} & 16        & 61.2  \\
      RotNet3D \cite{rotnet3d}           &  3D-ResNet18   & 16        & 62.9 \\
      ST-Puzzle \cite{stpuzzle}          &  3D-ResNet18   & 16        & 65.8  \\
      ClipOrder \cite{cliporder}         &  R(2+1)D-18\cite{r21d_v2}. &  -  & 72.4  \\
      DPC \cite{dpc}                     &  3D-ResNet34   & -         & 75.7  \\
      AoT \cite{arrow_of_time}           &  T-CAM         & -         & 79.4     \\
      SpeedNet \cite{speediness}         &  I3D\cite{kinetics}        & 64      & 81.1 \\
      VTHCL  \cite{vthcl}                &  3D-ResNet50  & 8         & 82.1 \\
      PacePrediction \cite{wang2020self} &  S3D-G\cite{r21d}          & 64    & 87.1  \\
      CoCLR  \cite{coclr}                &  S3D-G\cite{r21d}          & 32      & 87.6 \\\midrule
      Ours                            & 3D-ResNet50   & 8         & \textbf{88.9} \\
      \bottomrule
    \end{tabular}}
    \vspace{-15pt}
  \end{table*}

\begin{table}[t]
    \centering
    \caption{
        \textbf{Comparison across different pseudo-labeling schemes.}
        As there is no auxiliary network, FixMatch can only use the pseudo-labels generated by itself.
    }
    \label{tab-ablation-pseudo}
    \vspace{-5pt}
    \setlength{\tabcolsep}{15pt}
    \begin{tabular}{lc}
        \toprule
        
        Pseudo-Labeling & Top-1 \\ \midrule
        FixMatch      &  6.78 \\ \midrule
        Self-Confident        & 10.80   \\
        Opposite-Confident           & 11.13  \\
         Maximum             & 11.85  \\
        Average        & 12.16 \\ \midrule
        Cross           & 12.90  \\ 
        \bottomrule
    \end{tabular}
\end{table}

\noindent\textbf{Experimental Results.}
\cref{tab-ablation-pseudo} presents the results of different pseudo-labeling schemes.
The baseline strategy (FixMatch~\cite{sohn2020fixmatch}) performs the worst. 
Due to the lack of the auxiliary networks, the unlabeled data mainly distinguishable beyond the representation of the primary backbone rarely gets paired with confident pseudo labels since the scores of those unseen videos are easily below the threshold. 
After introducing the temporal information derived from the auxiliary network, clear improvements are observed. 
For the remaining strategies except the proposed cross-model scheme, there is a chance that the primary network will dominate the pseudo labeling decisions, leading to the wrong decisions for unlabeled samples. In contrast, for our cross-model strategy, each network always receives pseudo labels from its companion and never from itself, and this is shown to be more effective.

\section{Comparison to Self-Supervised Methods}  \label{sec-selfsupervision}
%
%
In this section, we compare CMPL with state-of-the-art self-supervised learning approaches.
We use UCF-101 as the labeled data and Kinetics-400 as the unlabeled data. 
A described in the main paper, CMPL jointly use labeled and unlabeled data in a semi-supervised manner.
As for self-supervised learning methods, we follow the standard protocol to use unlabeled data in Kinetics-400 for pre-training, followed by a fine-tuning on the labeled data in UCF-101.

As shown in \cref{table:sota}, in comparison to the CoCLR~\cite{coclr}, our model provides a performance gain of $1.3\%$ only with 8 frames input, indicating the effectiveness of CMPL.
It is a very encouraging result, suggesting that semi-supervised learning is a promising solution for action recognition with limited labeled data.
We hope that our result can provide a strong baseline for comparison with more self-supervised learning methods.

\section{Effects of Sampling Schemes}  \label{sec-sampling}
%
%
As illustrated in Section {\textcolor{red}{4.1}} of the main paper, the number of sampled videos is the same across different categories.
However, different from UCF-101, the distribution of videos across different categories is not balanced in Kinetics-400.
We re-sample a new video subset under the Kinetics-400 distribution, called `category-wise sampling scheme'.
To be specific, we first compute the number of each category and next randomly sample the videos from each category with the corresponding ratio and the total number.
\cref{tab-sampling} presents the results of different sampling schemes under the same setting of ablation study in Section {\textcolor{red}{4.3}} of the main paper.
Even with the unbalanced distribution, CMPL obtains nearly the same performance with the `uniform sampling' scheme, suggesting the robustness and generality of our approach.

\begin{table}[t]
    \setlength{\tabcolsep}{3.5pt}
        \caption{Study on sampling Scheme.}
        \label{tab-sampling}
        \centering
        \begin{tabular}{l|cc}
        \toprule
                            & Uniform(Default) & Category-Wise  \\ \hline
          Top-1            & 12.90            & 12.68        \\ 
    \bottomrule
    \end{tabular}
    \vspace{-10pt}
\end{table}

\section{3D-ResNet Network Structure} \label{sec-3dresnet}

\cref{tab-arch} shows the architecture of 3D-ResNet50.
It inherits the 2D-ResNet~\cite{resnet} and inflates the 2D kernel at $conv1$ across all stages.
The other convolution blocks are still in 2D format, focusing on the spatial semantics.
Moreover, there exist no temporal downsampling layers, in order to maintain long-temporal fidelity. 
Notably, we shrink the width of 3D-ResNet to a factor of 1/4 to use the 3D-ResNet50$\times$1/4 as the default auxiliary pathway.

\begin{table}[t]
    \centering
    \caption{\textbf{3D-ResNet50 Network Structure}.The  dimensions of convolution kernels are denoted by $\{K_T\times K_H \times K_W, K_C \}$ for temporal, height, width and channels sizes. The output size is in $\{C\times T \times S^2 \}$ format denoting channel, temporal and spatial size. We take input size of $3\times8\times224^2$ which utilizes 8 frames with 224 spatial resolution as an example.}
    \label{tab-arch}
    \vspace{-5pt}
    \small
    \begin{tabular}{ccc}
    \toprule
    Stage & Block  &   Output Size  \\ 
    \midrule
    \multirow{2}{*}{input} & \multirow{2}{*}{$-$} &  \multirow{2}{*}{$3\times8\times 224^2$}   \\
    & &  \\
    \midrule
    \multirow{2}{*}{conv$_1$} & \multicolumn{1}{c}{5$\times$7$\times$7, {64}} & \multirow{2}{*}{$64\times8\times 112^2$}  \\
    & stride 1, 2, 2 & \\
    \midrule
    \multirow{2}{*}{pool$_1$} & \multicolumn{1}{c}{1$\times$3$\times$3, max} &  \multirow{2}{*}{$64\times8\times 56^2$}  \\
    & stride 1, 2, 2 & \\
    \midrule
    \multirow{3}{*}{res$_2$} & \blocks{{256}}{{64}}{3} & \multirow{3}{*}{$256\times8\times 56^2$} \\
    & &  \\
    & &  \\
    \midrule
    \multirow{3}{*}{res$_3$} & \blocks{{512}}{{128}}{4}  &  \multirow{3}{*}{$512\times8\times 28^2$}   \\
    & &  \\
    & &  \\
    \midrule
    \multirow{3}{*}{res$_4$} & \blocks{{1024}}{{256}}{6} &  \multirow{3}{*}{$1024\times8\times 14^2$}  \\
    & &  \\
    & &  \\
    \midrule
    \multirow{3}{*}{res$_5$} & \blocks{{2048}}{{512}}{3} & \multirow{3}{*}{$2048\times8\times 7^2$}  \\
    & &  \\
    & & \\
    \bottomrule
  \end{tabular}
  \vspace{-5pt}
\end{table}

\end{document}